%% file: main.tex
\documentclass{article} % For LaTeX2e
\usepackage{aditi}
\renewcommand{\ForumContactRow}{%
  \begingroup\small\raggedright
    \ifx\ForumEmail\empty\else
      {\color{ForumAccent}\faEnvelope[regular]~}\ %
      \href{mailto:\ForumEmail}{\textcolor{ForumContactText}{\texttt{\ForumEmail}}}\par
      \vspace{\ForumContactGap}%
    \fi
    
  \endgroup
}
\usepackage{microtype}
\usepackage{xspace}
\usepackage{hyperref}
\usepackage{url}
\usepackage{booktabs}
\usepackage{multirow}
\usepackage{multicol}
\usepackage{lineno}
\usepackage{style} 
\usepackage{array}
\usepackage{hyperref}
\usepackage{fontawesome5} % 提供 \faGithub
\definecolor{skyblue}{RGB}{204,229,255}

\usepackage{algorithm}
\usepackage{algorithmic}

\usepackage{tabularx}
\newcolumntype{Y}{>{\centering\arraybackslash}X}

\usepackage[most]{tcolorbox}
\tcbset{
  lightbluebox/.style={
    colback=skyblue!70,        % light blue background
    colframe=skyblue!75!black, % dark blue frame
    boxrule=1pt,
    arc=4pt,
    auto outer arc
  }
}

\definecolor{darkblue}{rgb}{0, 0, 0.5}
\hypersetup{colorlinks=true, citecolor=darkblue, linkcolor=darkblue, urlcolor=darkblue}

\newcommand{\method}{\textsc{LenGuard-GPC}\xspace}

\setTitleruleGap{0.75pt}         % exact touch

\title{LenGuard-GPC: Length Guarding with Guided-Prompt Consistency for Spatial Reasoning Reinforcement Learning

}

\setauthors{Xingjian Tao$^{1}$ \authorsep Yiwei Wang$^{3}$\authorsep Yujun Cai$^{4}$ \authorsep Xinghua Qu$^{5}$  \authorsep Jing Tang$^{1,2}$\thanks{Corresponding Author: Jing Tang.}}
\setaffils{$^{1}$HKUST(GZ),\quad $^{2}$HKUST, \quad $^{3}$UC Merced, \quad $^{4}$University of Queensland, \quad $^{5}$Independent Researcher}

\setemail{taoxj2001@outlook.com}
\setemails{\href{https://github.com/taoxj2001/LenGuard-GPC}{\faGithub\ https://github.com/taoxj2001/LenGuard-GPC}}

\usepackage{xcolor}
\usepackage{minitoc}
\usepackage{graphicx}
\definecolor{jcg}{RGB}{100,160,0}
\definecolor{sachin}{RGB}{0,0,150}
\definecolor{hqz}{RGB}{160,100,100}
\definecolor{gnz}{HTML}{64B5F6}
\usepackage[most]{tcolorbox} % 正确的方式
\tcbuselibrary{skins,breakable} % 确保加载了这些库以获得更强大的功能和样式

\definecolor{myDarkGreen}{RGB}{50, 70, 70} % 深灰绿色，用于标题背景
\definecolor{myLightGray}{RGB}{240, 240, 240} % 浅灰色，用于内容背景

\definecolor{titlebgcolor}{RGB}{70, 80, 100}
\definecolor{bodybgcolor}{RGB}{245, 245, 245}
\definecolor{bordercolor}{RGB}{120, 120, 120}
\definecolor{darkblue}{rgb}{0.0, 0.0, 0.55}   % 定义深蓝色
\definecolor{darkgreen}{rgb}{0.0, 0.5, 0.0}   % 定义深绿色
\definecolor{darkred}{rgb}{0.6, 0.0, 0.0}     % 定义深红色

\usepackage{url}
\usepackage{amsthm}
\usepackage{minitoc}
\usepackage{enumitem}
\usepackage{todonotes}
\usepackage{float}
\usepackage{listings}
\usepackage{caption}
\usepackage{cleveref}
\usepackage[table]{xcolor} % 推荐方式，table选项会自动加载colortbl
\usepackage{graphicx}

\usepackage{booktabs}

\definecolor{myLightBlue}{RGB}{230, 240, 255} 

\usepackage[T1]{fontenc}
\newtcolorbox{responsebox}[2][]{
    breakable,
    enhanced,
    colback=white,             % 背景色：白色
    colframe=blue!50!black,    % 边框色：深蓝色
    coltext=black,             % 文本色：黑色
    coltitle=white,    % 标题颜色：深蓝色
    fonttitle=\bfseries\rmfamily, % 标题字体：粗体、衬线
    arc=3mm,                   % 圆角
    boxrule=1pt,
    title=#2,
    #1
}
\definecolor{lightblue}{RGB}{235,243,252}

\definecolor{mybgcolor}{RGB}{235, 235, 250}
\definecolor{myGreen}{RGB}{240, 250, 240}

\newtcolorbox{takeawaybox}[1][]{
  enhanced,
  colback=mybgcolor, % 设置背景颜色
  colframe=black,    % 设置边框颜色
  boxrule=0.5pt,     % 边框线宽
  arc=3mm,           % 圆角半径

  attach boxed title to top left={yshift=-0.25em, xshift=1em},
  fonttitle=\bfseries, % 标题字体加粗
  title={#1},          % 将环境的参数作为标题
  boxed title style={
    colback=black,     % 标题框背景为黑色
    sharp corners,     % 标题框使用直角
  },
}
\newtcolorbox{equationbox}[1]{
  colback=white,                % 盒子主体内容的背景色 (白色)
  colframe=gray!75!black,       % 边框颜色 (深灰色)
  boxrule=1pt,                  % 边框粗细
  
  title=#1,                     % #1 表示 \begin{...} 后的参数，即标题文字
  attach boxed title to top left={yoffset=-2mm, xshift=2mm}, % 标题栏位置
  
  colbacktitle=gray!75!black,   % 标题栏背景色 (深灰色)
  coltitle=white,               % 标题栏文字颜色 (白色)
  fonttitle=\bfseries\sffamily, % 标题栏字体 (加粗, 无衬线)
  
  boxed title style={
    boxrule=0pt,                % 标题栏本身的边框 (设为0)
    frame code={}               % 移除标题栏的额外边框
  }
}

\begin{document}

\maketitle

\begin{abstract}
Multi-view spatial reasoning requires vision-language models to compare visual evidence across images, align object correspondences, and infer spatial relations over long visual contexts, a setting where chain-of-thought reasoning tends to grow verbose without becoming more accurate. Reinforcement learning with verifiable rewards is a natural fit for this task, but standard GRPO reward relies on sparse outcome-level feedback and gives no signal about where a reasoning trajectory goes wrong, nor any control over its length. We propose \method, a dense reward framework that addresses both problems together. For each sampled trajectory, it compares the token-wise predictive distributions under a standard prompt and a guided prompt, and uses the resulting token-sum KL divergence as a dense reward signal. Since this KL penalty accumulates over tokens and would otherwise reward shorter responses regardless of their quality, we introduce a staged length bonus that keeps reasoning length within a controlled range without simply encouraging brevity. On six multi-view spatial reasoning benchmarks, \method improves accuracy over vanilla GRPO while reducing average response length.

\end{abstract}

\input{1intro}

 \input{2related_work}

\input{3method}

 \input{4experiments}

\input{5conclusion}

 % \input{entropy_effect}
 % \input{theory}
 % \input{conclusion}

% \section*{Acknowledgements}
\newpage
\bibliography{iclr2026_conference}
\bibliographystyle{colm2025_conference}

% \newpage
% \appendix
% \renewcommand \thepart{} % make "Part" text invisible
%     \renewcommand \partname{}
% \part{Appendix} % Start the appendix part
% \input{appendix}
    \parttoc % Insert the appendix TOC
\end{document}

%% file: 1intro.tex
\section{Introduction}

 \begin{figure}[t!]
    \centering

        \includegraphics[width=1\linewidth]{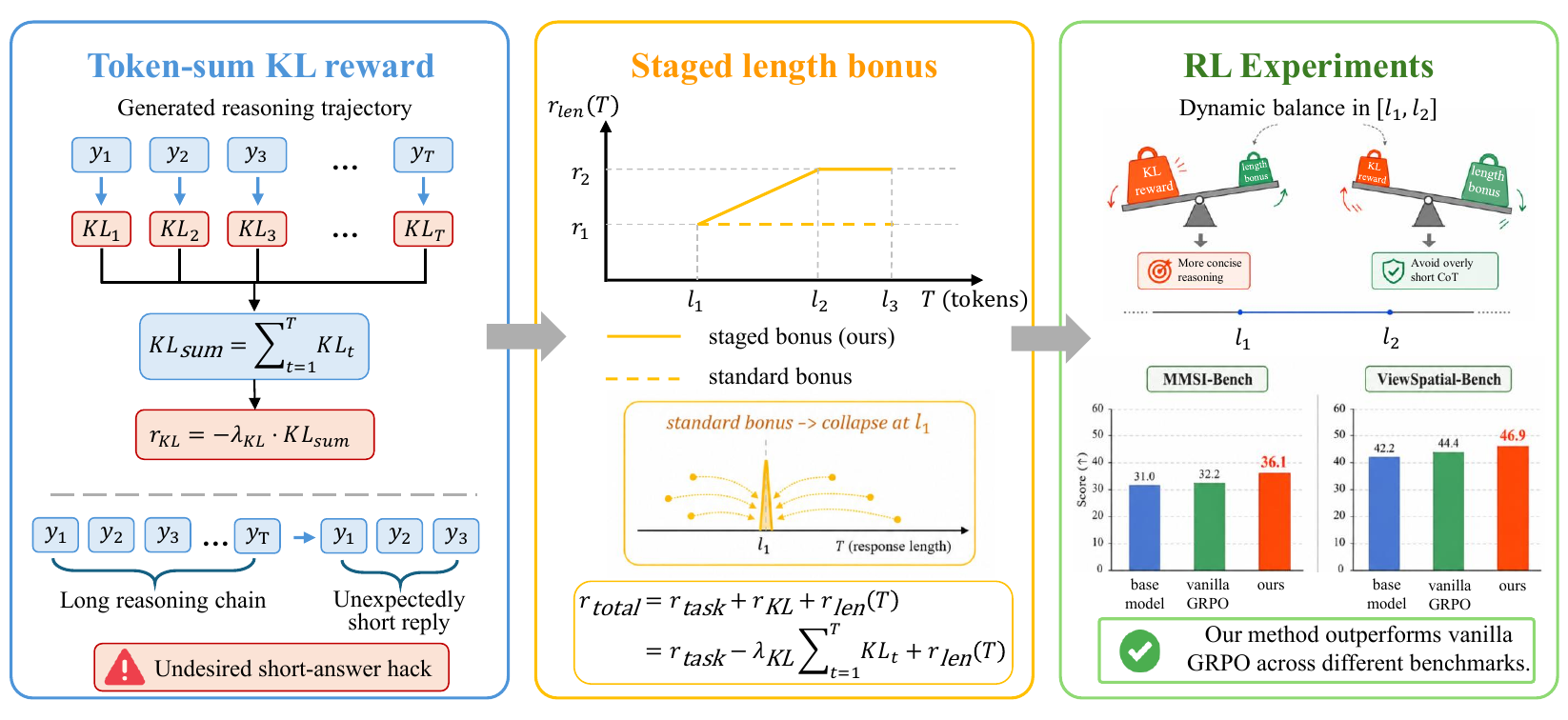}

        \caption{Overview of \method for multi-view spatial reasoning. Within the staged interval $[l_1,l_2]$, \method adaptively balances the negative KL-based consistency reward and the length bonus. When the mean per-token KL is high, the consistency reward dominates and encourages more concise reasoning. When the mean per-token KL is low, the length bonus becomes relatively stronger, preventing the model from collapsing into overly short and content-poor chains of thought.}

        \label{fig:pkpo_entropy}
        
\end{figure}

Vision-language models (VLMs) have recently made remarkable progress in bridging visual perception and language-based reasoning. By scaling multimodal pretraining and instruction tuning, models such as LLaVA~\citep{liu2023visual, liu2024improved, li2024llava}, Flamingo~\citep{alayrac2022flamingo}, Gemini~\citep{team2023gemini}, and Qwen-VL~\citep{bai2023qwen,wang2024qwen2,bai2025qwen3vltechnicalreport} have achieved strong performance on a wide range of vision-language tasks such as visual question answering, image captioning and visual grounding~\citep{kan2025taco, sarto2025image, wang2025vgr,wu2025postalign,tao2026mitigating,gou2025uground,tao2025understanding}, where a single image usually contains enough evidence to support an answer. 
Multi-view spatial reasoning is a different kind of problem: given several images of the same scene taken from different viewpoints, a model has to relate objects across images before it can answer a question about their spatial relations.

Consider two images of the same room, one from the front and one from the side. A model must recognize that a sofa near the wall in one image and a sofa near the door in the other are the same object before it can answer a question such as which object is closer to the window, a step that standard vision-language benchmarks never require and that even strong VLMs tend to fail at. When prompted to reason step by step over several such images, models typically describe what they see in each image in turn, producing traces that grow longer without ever explicitly resolving which object in one image corresponds to which in another, the step the question actually depends on. Longer traces can also compound errors: once several claims accumulate across views, an early misreading tends to persist uncorrected, and irrelevant detail crowds out the one relation that matters. Reasoning quality here is therefore not a function of length; what matters is whether the trace resolves cross-view correspondence at all.

Reinforcement learning with verifiable rewards offers a promising direction for improving reasoning behavior, since spatial reasoning tasks often admit objective answer-level supervision. Methods such as GRPO can optimize models using outcome rewards without requiring dense human annotations. However, directly applying reward-only optimization to multi-view spatial reasoning remains insufficient. A final correctness reward provides limited information about where the reasoning process goes wrong, especially when the failure is caused by subtle local errors such as mismatched viewpoints, incorrect object correspondence, or invalid spatial relation inference. Moreover, outcome-only reward optimization may encourage undesirable reasoning patterns, including unnecessarily long rationales, repetitive visual descriptions, or superficial answer guessing. Therefore, an effective RL objective for multi-view spatial reasoning should provide denser process-level guidance while also controlling the length and structure of the reasoning trajectory.

To address these challenges, we propose \method, a length-guarded guided-prompt consistency framework for GRPO-based spatial reasoning. The guided-prompt design is inspired by Self On-Policy Distillation~\citep{hubotter2026reinforcement}, which compares model behavior under different prompting conditions. The central idea is to use a guided prompt as a reference condition for evaluating the model's sampled trajectory. Given a generated response, we compute the token-wise predictive distributions under two conditions: the original prompt and a guided prompt that provides additional reasoning guidance. We then measure the KL divergence between these two distributions along the trajectory. This guided-prompt consistency score reflects the consistency between the model's current reasoning behavior and the behavior encouraged by the guided prompt. Rather than introducing an auxiliary supervised objective, \method directly incorporates the trajectory-level KL score into the reward function used by GRPO. In this way, the model receives denser feedback than a final correctness reward while preserving the simplicity and flexibility of reinforcement learning optimization.

However, using token-sum KL as a reward component introduces a length-related challenge. Since KL scores accumulate over tokens, longer reasoning chains naturally produce larger cumulative values. This may create an incentive for the model to generate overly short responses, even when a moderate amount of reasoning is necessary. A simple flat length bonus can partially compensate for this bias, but it may cause response lengths to concentrate around the earliest rewarded boundary. To avoid both short-answer collapse and boundary-seeking behavior, \method introduces a staged length bonus controlled by length boundaries $(l_1,l_2,l_3)$ and bonus levels $(r_1,r_2)$. The bonus is not applied before $l_1$, increases from $r_1$ to $r_2$ within the interval $[l_1,l_2]$, and remains constant until $l_3$. This schedule creates a dynamic balance between KL-guided compactness and length-based sufficiency: when the mean per-token KL is high, the consistency reward encourages more concise reasoning; when the mean per-token KL is low, the length bonus prevents the model from collapsing into overly short and content-poor chains of thought.

Although \method is developed for multi-view spatial reasoning, its principle can extend to broader long-context multimodal reasoning tasks, such as video understanding, multi-image comparison, document-grounded visual reasoning, and embodied perception. These tasks also require models to reason over dense multimodal inputs, where sparse outcome rewards may be insufficient and unconstrained chain-of-thought generation can become verbose or poorly grounded. By requiring only a guided prompt and token-level distributional comparison, \method offers a general reward-based framework for improving reasoning behavior in multimodal reinforcement learning.

Our contributions are summarized as follows. First, we identify a key optimization challenge in GRPO-based multi-view spatial reasoning: sparse outcome rewards provide limited guidance for local cross-view reasoning errors, while naive token-sum KL reward shaping can bias models toward overly short responses. Second, we propose \method, which leverages guided-prompt token-wise KL consistency scores as a GRPO-compatible reward signal. Third, we introduce a staged length-guard mechanism that mitigates both short-answer collapse and boundary-seeking behavior. Finally, we show that \method provides a general RL framework for utilizing guided-prompt consistency rewards in long-context multimodal reasoning.

\begin{tcolorbox}[
  enhanced, breakable,
  colframe=black!12, boxrule=0.35pt, arc=1mm,
  title={\textbf{Summary of Our Main Contribution}},
  coltitle=black, fonttitle=\sffamily\bfseries,
  colbacktitle=blue!15!white,
  colback=blue!5!white,
  boxed title style={
    sharp corners, boxrule=0pt,
    top=3pt, bottom=3pt, left=4mm, right=4mm,
    borderline={0.5pt}{0pt}{black!10}
  },
  attach boxed title to top left={xshift=4mm,yshift*=-1.2mm},
  boxsep=1.5mm, top=1.5mm, bottom=1.5mm, left=4mm, right=4mm,
  before skip=10pt, after skip=10pt
]
\begin{enumerate}[topsep=0pt,leftmargin=10pt]\setlength{\itemsep}{0pt}
\item We identify a key challenge in GRPO-based multi-view spatial reasoning: sparse outcome rewards offer limited guidance for local cross-view reasoning errors, while naive token-sum KL shaping can bias models toward overly short responses.
\item We introduce \method, a length-guarded guided-prompt consistency reward that converts guided-prompt token-wise distributional discrepancies into a GRPO-compatible reward signal, enabling dense process-level feedback without an additional supervised loss.
\item We propose a staged length-guard mechanism that mitigates short-answer collapse and boundary-seeking behavior, while demonstrating \method as a general reward-based framework for long-context multimodal reinforcement learning.
\end{enumerate}
\end{tcolorbox}

%% file: 2related_work.tex
\section{Related Work}\label{sec:related work}

\subsection{Reinforcement Learning for Multimodal Large Language Model Reasoning}

Reinforcement learning (RL) and preference optimization have been increasingly used to improve the reasoning quality and behavioral alignment of multimodal large language models (MLLMs)~\citep{sun2024aligning, yu2024rlhf, yu2024rlaif, xie2024v,huang2025vision}. Recent reasoning-oriented RL methods optimize models with verifiable rewards or preference signals to encourage stronger multimodal deliberation. For example, Insight-V~\citep{dong2025insight} learns from self-generated reasoning trajectories selected by a multi-agent framework, while R1-VL~\citep{zhang2025r1} adopts step-wise GRPO with dense rule-based rewards. GRPO-style training has also been extended to video reasoning, including Video-R1~\citep{feng2025video}, Video-RTS~\citep{wang2025video}, and Video-STR~\citep{wang2025video2}. In parallel, several works encourage models to ``observe first'' through explicit context descriptions or observation stages, such as HumanOmniV2~\citep{yang2025humanomniv2}, Visionary-R1~\citep{xia2025visionary}, ViewFusion~\citep{tao2026viewfusion} and Observe-R1~\citep{guo2025observe}.
Despite these advances, existing RL-based MLLM reasoning methods pay limited attention to reasoning length. In long-context multimodal tasks, overly long chains may introduce redundant or inconsistent reasoning, while overly short responses may be insufficient. Our work addresses this issue with a teacher-prompt KL reward and a staged length guard to balance reasoning sufficiency and compactness.

\subsection{Spatial reasoning with MLLMs}
Spatial reasoning has emerged as a key frontier for MLLMs, aiming to move beyond object recognition toward understanding relative position, orientation, viewpoint transformation, and occlusion-aware relations in 3D scenes. A growing body of work seeks to strengthen spatial reasoning in MLLMs by improving grounding and spatial representations, for example via spatially aware instruction tuning and curated supervision~\citep{liu2023visual2, tong2024cambrian, chen2024spatialvlm,yu2025far,gholami2025spatial,wu2025spatialscore,zhao2025spacemind, batra2025spatialthinker, wang2025visioncube, fan2025vlm, li2024topviewrs}. More recently, Visual Spatial Tuning~\citep{yang2025visual} trains vision-language models with large-scale spatial perception and reasoning data, producing notably stronger spatial reasoning performance and improved generalization across spatial benchmarks. 

To systematically evaluate these advances under multi-view inputs, several benchmarks have been proposed~\citep{zhang2025sphere, lee2025spatialmosaic}. MMSI-Bench~\citep{yang2025mmsi} focuses on multi-view, multi-image spatial intelligence and includes problems that require aligning evidence across views rather than solving from a single snapshot. ViewSpatial~\citep{li2025viewspatial} further stresses viewpoint-dependent spatial localization and cross-view reference frames, revealing substantial generalization gaps when camera viewpoints shift.

%% file: 3method.tex
\section{\method}

\subsection{Reasoning-Length Challenge in Multi-View Spatial Reasoning}

Chain-of-thought prompting is widely used to elicit intermediate reasoning in language and multimodal models~\citep{wei2022chain,zhang2023multimodal}. However, in multi-view spatial reasoning, long reasoning traces often become verbose and inefficient, repeatedly describing irrelevant details or comparing similar regions without advancing the target inference, which distracts from key cross-view evidence.
Recent studies suggest that excessive chain-of-thought length can hurt both accuracy and computational efficiency, while concise reasoning may achieve comparable or better performance~\citep{nayab2024concise,xu2025chain,hassid2025don}. Thus, a key challenge is to regulate reasoning length and encourage concise, sufficient, and visually grounded reasoning.

Simply encouraging shorter reasoning is insufficient. Although long-to-short methods can improve efficiency, overly aggressive reduction may reduce reasoning transparency or cause behavioral inconsistency~\citep{yang2025long}. Likewise, rewards favoring the shortest valid output may lead to answer-only behavior, format collapse, or other reward-hacking strategies rather than genuinely improved reasoning.
These observations motivate our RL formulation. We introduce a guided-prompt consistency reward for dense trajectory-level feedback and combine it with a staged length guard to avoid both overly long and overly short reasoning. Inspired by Self On-policy Distillation~\citep{hubotter2026reinforcement}, our method compares behavior under different prompting conditions, but uses this comparison as an RL reward rather than a supervised distillation objective.

\subsection{Guided-Prompt Consistency Reward}

Given a multi-view reasoning instance, we denote the visual inputs as $\mathcal{I}=\{I_1,\ldots,I_m\}$ and the question as $q$. The standard prompt is written as $x=(\mathcal{I},q)$, under which the policy model $\pi_\theta$ samples a reasoning trajectory $y=(y_1,\ldots,y_T)$. To provide denser feedback during training, we additionally construct a guided prompt $x^{\mathrm{gp}}=(\mathcal{I},q,h)$, where $h$ provides task-specific guidance for evaluating the generated trajectory. In our implementation, $h$ is a one-sentence guidance generated by GPT-5.3-Chat for each training instance. The guided prompt is used only for reward computation during training and is not required at inference time.

For the same sampled trajectory $y$, we compare the model's token-level predictive distributions under the standard prompt and the guided prompt. Both distributions are computed along the same generated trajectory using teacher forcing:
\begin{equation}
p_t^{\mathrm{std}}=\pi_\theta(\cdot \mid x,y_{<t}), \quad
p_t^{\mathrm{gp}}=\pi_\theta(\cdot \mid x^{\mathrm{gp}},y_{<t}), \quad
D_t=D_{\mathrm{KL}}\!\left(p_t^{\mathrm{std}} \,\|\, p_t^{\mathrm{gp}}\right).
\end{equation}
Here, $D_t$ measures how much the model's prediction under the standard prompt deviates from its prediction under the guided-prompt condition at the same reasoning state.

We aggregate the token-level KL scores over the full trajectory and define the guided-prompt consistency reward as the negative accumulated divergence:
\begin{equation}
S_{\mathrm{KL}}(y)=\sum_{t=1}^{T}D_t, \quad
r_{\mathrm{KL}}(y)=-S_{\mathrm{KL}}(y).
\end{equation}
In this way, trajectories that remain closer to the guided-prompt behavior receive a smaller penalty, while trajectories that deviate substantially from the guided condition receive a lower reward. Importantly, this KL score is used as an RL reward rather than an auxiliary supervised objective. Thus, the model is not forced to imitate a fixed reasoning path, but is instead encouraged to generate trajectories that are more consistent with the guided-prompt evaluation signal.

\subsection{Staged Length Guard}

Although the guided-prompt consistency reward provides dense trajectory-level feedback, it also introduces a length-related bias. Since the KL score is accumulated over tokens, longer reasoning trajectories tend to receive larger KL penalties even when their reasoning process is valid. If used alone, this penalty may encourage the policy to shorten its responses excessively, resulting in insufficient or answer-only reasoning. A straightforward solution is to add a flat length bonus after a minimum length threshold, but this can create another degenerate behavior: the model may concentrate its response length around the earliest rewarded boundary, producing just-enough but underdeveloped reasoning.

To address this issue, we introduce a staged length guard. Instead of assigning a constant length bonus, we divide the reasoning length into stages. No length bonus is assigned before the lower boundary $l_1$, which discourages overly short responses. Within the interval $[l_1,l_2]$, the bonus gradually increases from $r_1$ to $r_2$, preventing the policy from collapsing all responses around $l_1$. After $l_2$, the bonus is kept constant, so the model does not receive additional incentives for producing unnecessarily long reasoning chains. Formally, the staged length bonus is defined as
\begin{equation}
r_{\mathrm{len}}(T)=
\begin{cases}
0, & T < l_1 \ \text{or}\ T > l_3,\\
r_1 + (r_2-r_1)\dfrac{T-l_1}{l_2-l_1}, & l_1 \leq T \leq l_2,\\
r_2, & l_2 < T \leq l_3.
\end{cases}
\end{equation}
where $T$ denotes the number of generated tokens, $(l_1,l_2,l_3)$ are the length boundaries, and $(r_1,r_2)$ are the lower and upper bonus levels. In practice, $l_3$ corresponds to the maximum effective reasoning length considered by the length guard.

The staged design balances compactness and sufficiency during RL optimization. When the mean per-token KL is high, the KL reward discourages unnecessary or misaligned reasoning steps. When the mean per-token KL is low, the length guard prevents the model from collapsing into overly short responses. As a result, the policy is encouraged to produce reasoning trajectories that are not merely short, but concise, sufficient, and better aligned with the guided-prompt evaluation signal.

\subsection{GRPO Training with Guided-Prompt Consistency Reward}

\subsubsection{Preliminary: Group Relative Policy Optimization}

Group Relative Policy Optimization (GRPO) optimizes a policy by comparing multiple responses sampled from the same input. Given a multi-view reasoning prompt $x$, the policy $\pi_\theta$ samples $G$ trajectories $\{y_i\}_{i=1}^{G}$, where each $y_i=(y_{i,1},\ldots,y_{i,T_i})$ contains the reasoning process and final answer. GRPO estimates trajectory advantages through group-level reward normalization without an external value model.

For each sampled trajectory, the task reward is a weighted sum of accuracy and format rewards:
\begin{equation}
r_{\mathrm{task}}(y_i)
=
\lambda_{\mathrm{acc}} r_{\mathrm{acc}}(y_i)
+
\lambda_{\mathrm{format}} r_{\mathrm{format}}(y_i).
\end{equation}
Here, $r_{\mathrm{acc}}(y_i)$ evaluates answer correctness, while $r_{\mathrm{format}}(y_i)$ encourages valid output formatting.

The final trajectory reward combines the task reward, guided-prompt consistency reward, and staged length bonus:
\begin{equation}
r_i
=
r(y_i)
=
\lambda_{\mathrm{task}} r_{\mathrm{task}}(y_i)
+
\lambda_{\mathrm{KL}} r_{\mathrm{KL}}(y_i)
+
\lambda_{\mathrm{len}} r_{\mathrm{len}}(T_i).
\end{equation}
Here, $r_{\mathrm{KL}}(y_i)$ measures token-wise deviation between the standard and guided prompts, while $r_{\mathrm{len}}(T_i)$ regulates reasoning length.

GRPO then normalizes the trajectory rewards within each group to obtain relative advantages:
\begin{equation}
A_i=
\frac{
r_i-\mathrm{mean}\left(\{r_j\}_{j=1}^{G}\right)
}{
\mathrm{std}\left(\{r_j\}_{j=1}^{G}\right)+\epsilon
}.
\end{equation}
where $\epsilon$ is a small constant for numerical stability. The standard GRPO objective then encourages above-average trajectories and suppresses below-average ones.

During training, responses are sampled only from the standard prompt $x$, while the guided prompt $x^{\mathrm{gp}}$ is used solely to compute the guided-prompt consistency reward. Thus, our method introduces no additional inference-time overhead.

\subsubsection{Training Procedure}
Algorithm~\ref{alg} summarizes the training process. For each instance, we construct both the standard prompt and the guided prompt. The policy samples a group of trajectories from the standard prompt, and each trajectory is evaluated using the accuracy reward, format reward, guided-prompt consistency reward, and staged length bonus. The combined rewards are normalized within the group and used to update the policy with GRPO.

Overall, the proposed training procedure treats guided-prompt consistency as an RL reward rather than a target sequence or an auxiliary loss. By combining task correctness, format control, guided-prompt consistency, and staged length regulation, the model is optimized to produce reasoning trajectories that are correct, compact, and sufficiently grounded.

\begin{algorithm}[t]
\caption{GRPO Training with Guided-Prompt Consistency Reward}
\label{alg}
\begin{algorithmic}[1]
\REQUIRE Training instance $(\mathcal{I},q,a)$, policy $\pi_\theta$, group size $G$
\STATE Construct the standard prompt $x=(\mathcal{I},q)$ and the guided prompt $x^{\mathrm{gp}}=(\mathcal{I},q,h)$.
\STATE Sample $G$ trajectories $\{y_i\}_{i=1}^{G}$ from $\pi_\theta(\cdot \mid x)$.
\FOR{each trajectory $y_i$}
    \STATE Compute reward components $r_{\mathrm{acc}}$, $r_{\mathrm{format}}$, $r_{\mathrm{KL}}$, and $r_{\mathrm{len}}$.
    \STATE Combine them into the total reward $r_i=r(y_i)$.
\ENDFOR
\STATE Normalize $\{r_i\}_{i=1}^{G}$ within the group to obtain advantages $\{A_i\}_{i=1}^{G}$.
\STATE Update $\pi_\theta$ with the standard GRPO objective.
\end{algorithmic}
\end{algorithm}

%% file: 4experiments.tex
\section{Experiments}
\label{sec:experiments}

\subsection{Experimental Setup}
\label{sec:exp_setup}

\paragraph{Training Data.}
Our training set is derived from VST-500K~\citep{yang2025visual}. We sample approximately 13,000 instances from its multi-view spatial reasoning data, where each example contains multiple images captured from different viewpoints and requires the model to reason over cross-view spatial relations. Unlike single-image visual question answering, these samples require the model to compare visual evidence across views, establish object correspondences, and infer spatial layouts from partially overlapping observations. Therefore, this training set provides a suitable foundation for studying GRPO-based multi-view spatial reasoning and for evaluating whether guided-prompt consistency rewards can improve reasoning behavior under long and visually dense contexts.

\paragraph{Implementation Details.}
We use Qwen3-VL-8B-Instruct as the base model and train the policy with GRPO. The learning rate is set to $1\times10^{-6}$. The task reward is composed of an accuracy reward and a format reward:
\begin{equation}
r_{\mathrm{task}}(y)
=
\lambda_{\mathrm{acc}} r_{\mathrm{acc}}(y)
+
\lambda_{\mathrm{format}} r_{\mathrm{format}}(y).
\end{equation}
Here, $r_{\mathrm{acc}}(y)$ measures whether the final answer is correct, and $r_{\mathrm{format}}(y)$ measures whether the response follows the required output format. In implementation, we use a format reward of $0.2$ to encourage valid response formatting.

The guided-prompt consistency reward is computed from the accumulated token-level KL score:
\begin{equation}
S_{\mathrm{KL}}(y)=\sum_{t=1}^{T}D_t,
\end{equation}
and is incorporated into the final reward with coefficient $\lambda_{\mathrm{KL}}=5\times10^{-3}$. Equivalently, this corresponds to a KL-score coefficient of $-5\times10^{-3}$ in implementation.

For the staged length guard, we set the length boundaries as $l_1=160$, $l_2=360$, and $l_3=768$. The length bonus starts from $r_1=0.25$ at $l_1=160$ tokens and increases by $2.5\times10^{-4}$ for each additional token until it reaches $r_2=0.30$ at $l_2=360$ tokens. From $l_2$ to $l_3$, the length bonus remains fixed at $0.30$. No length bonus is granted beyond $l_3=768$ tokens, so the staged guard does not encourage excessively long reasoning trajectories. 

The final reward combines task performance, guided-prompt consistency, and staged length regulation:
\begin{equation}
r(y)
=
\lambda_{\mathrm{task}} r_{\mathrm{task}}(y)
-
\lambda_{\mathrm{KL}} S_{\mathrm{KL}}(y)
+
\lambda_{\mathrm{len}} r_{\mathrm{len}}(T).
\end{equation}
These settings are designed to balance the negative guided-prompt consistency reward with sufficient reasoning length, preventing short-answer collapse while avoiding unnecessary verbosity.

\paragraph{Evaluation Setting.}
We evaluate the trained models on six benchmarks covering both in-domain and out-of-domain reasoning scenarios. Specifically, MMSI-Bench~\citep{yang2025mmsi}, MindCube, and ViewSpatial-Bench~\citep{li2025viewspatial} are multi-image spatial reasoning benchmarks, where each instance requires the model to compare visual evidence across multiple views and infer spatial relations. These benchmarks are most aligned with our training data, which also consists of multi-view spatial reasoning samples. Notably, MindCube is used only for evaluation, and we do not use its provided training set.

To further assess the generalization ability of our method and verify that the training procedure does not overfit to the in-domain data, we additionally evaluate on three out-of-domain benchmarks: 3DSR~\citep{ma20253dsrbench} and CV-Bench~\citep{tong2024cambrian}, which focus on single-image spatial reasoning, and BLINK~\citep{fu2024blink}, which involves multi-image reasoning but is not primarily centered on spatial relation inference. This evaluation protocol allows us to examine whether the proposed guided-prompt consistency reward improves performance on in-domain multi-view spatial reasoning tasks while maintaining robustness on related spatial and multi-image reasoning tasks beyond the training distribution.

\input{tabs/main_result}

\subsection{Quantitative Results}
\label{sec:quant_results}

Table~\ref{tab:main_results} reports the overall accuracy on six evaluation benchmarks. Compared with the base Qwen3-VL-8B-Instruct model, vanilla GRPO improves the average accuracy from $51.4\%$ to $55.3\%$, showing that RL training can enhance multi-view spatial reasoning performance. Our \method further improves the average accuracy to $57.3\%$, yielding a gain of $+2.0$ points over vanilla GRPO and $+5.9$ points over the base model.

The improvement is most pronounced on benchmarks that follow the same multi-view spatial reasoning domain as our training data. On MMSI, \method improves over vanilla GRPO from $32.2\%$ to $36.1\%$, corresponding to a $+3.9$ point gain. On MindCube, which is used only for evaluation without using its provided training set, the accuracy increases from $44.7\%$ to $48.4\%$. On ViewSpatial, \method further improves the accuracy from $44.4\%$ to $46.9\%$. These results suggest that the proposed reward design generalizes beyond the exact training source while remaining effective within the multi-view spatial reasoning domain.

On out-of-domain benchmarks, \method also maintains consistent gains over vanilla GRPO. It improves 3DSR from $59.4\%$ to $60.1\%$, BLINK from $65.7\%$ to $66.8\%$, and CV-Bench from $85.2\%$ to $85.5\%$. Although these gains are smaller than those on in-domain benchmarks, they indicate that the guided-prompt consistency reward and staged length guard do not overfit to the training distribution or sacrifice general reasoning ability.

Compared with existing open-source baselines, \method achieves the best average performance among the listed open-source models. These results demonstrate that explicitly regulating reasoning behavior with guided-prompt consistency and staged length control is effective for improving multi-view spatial reasoning while preserving robustness on out-of-domain evaluations.

Table~\ref{tab:finegrained_mmsi} further provides a fine-grained breakdown on MMSI-Bench. Compared with vanilla GRPO, \method improves the overall accuracy from $32.2\%$ to $36.1\%$ and achieves gains in most subcategories. The largest improvements appear in cross-view positional reasoning, including Cam.--Obj. ($+13.9$ points), Cam.--Reg. ($+12.0$ points), Obj.--Reg. ($+7.1$ points), and Obj.--Obj. ($+5.3$ points), suggesting that our reward design mainly strengthens spatial relation reasoning across views. In addition, \method improves Attribute Meas., Motion Cam., and MSR, while the drops on Attribute Appr. and Motion Obj. indicate that appearance-based attributes and object-motion reasoning remain challenging.

Although \method evaluates each sampled trajectory under both the standard and guided prompts, this does not require two full autoregressive generations. The trajectory is sampled only once from the standard prompt, while the guided-prompt distribution is computed over the fixed trajectory using teacher forcing, allowing token-level predictions to be evaluated in parallel. Moreover, the remaining GRPO operations, including advantage computation, backpropagation, and parameter updates, are shared with the vanilla training pipeline. Consequently, the additional guided-prompt evaluation increases the overall training time by only approximately $1.6\times$, rather than doubling it. The shorter trajectories induced by \method further offset part of this additional cost.uality here is therefore not a function of length; what matters is whether the trace resolves cross-view correspondence at all.
\input{tabs/result_qwen3}

\subsection{Ablation Study}
\label{sec:ablation}

We conduct ablation studies on MMSI-Bench to evaluate the contributions of guided-prompt consistency and the staged length reward. As shown in~\Cref{tab:ablation}, vanilla GRPO improves the base Qwen3-VL-8B-Instruct model from 31.1\% to 32.2\%. Adding guided-prompt consistency without the staged length reward further improves the accuracy to 33.5\%, showing that the proposed consistency reward provides additional supervision beyond the standard GRPO objective.
The full \method achieves 36.1\%, outperforming the variant without the staged length reward by 2.6 points. Improvements are observed across most fine-grained categories, especially cross-view positional reasoning. These results indicate that the staged length reward effectively complements guided-prompt consistency by preventing overly short reasoning while preserving the benefits of compact reasoning.

\vspace{40pt}
\subsection{Token Efficiency Analysis}
\label{sec:token_efficiency}

\begin{wrapfigure}{r}{0.66\textwidth}
\vspace{-1.0em}
\centering
\includegraphics[width=0.64\textwidth]{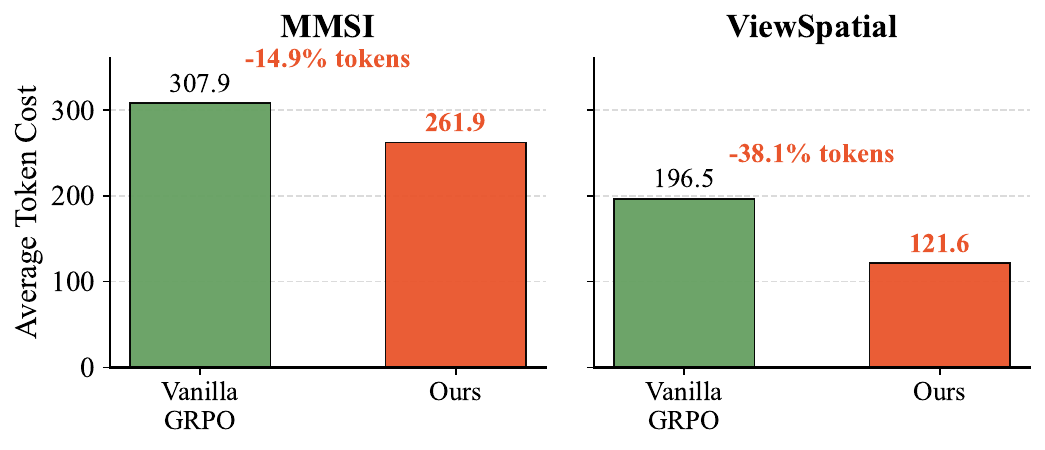}
\vspace{-0.8em}
\caption{Average generated token cost on MMSI and ViewSpatial. Compared with vanilla GRPO, \method produces shorter reasoning trajectories while maintaining stronger accuracy, indicating more compact and adaptive chain-of-thought reasoning.}
\label{fig:token_cost}
\vspace{-1.0em}
\end{wrapfigure}

In addition to accuracy, we further evaluate whether the proposed method can produce more compact reasoning trajectories. Figure~\ref{fig:token_cost} compares the average generated token cost between vanilla GRPO and \method on MMSI and ViewSpatial. Compared with vanilla GRPO, \method reduces the average token cost from $307.9$ to $261.9$ on MMSI, corresponding to a $14.9\%$ reduction. The reduction is more substantial on ViewSpatial, where the average token cost decreases from $196.5$ to $121.6$, corresponding to a $38.1\%$ reduction.

This result suggests that \method does not simply encourage uniformly long or short responses, but instead makes the reasoning length more adaptive. ViewSpatial contains both single-image and multi-image questions with different levels of reasoning difficulty. For easier single-image cases, the model can avoid unnecessary chain-of-thought expansion, while for harder multi-view cases, it can still retain sufficient reasoning when needed. Therefore, the proposed guided-prompt consistency reward and staged length guard compress redundant reasoning steps and improve the flexibility of reasoning length, leading to more efficient multimodal spatial reasoning.

\input{tabs/ablation}

%% file: tabs/main_result.tex
\begin{table*}[t]
\centering
\fontsize{8pt}{9.6pt}\selectfont
\setlength{\tabcolsep}{2.2pt}
\begin{tabularx}{1\linewidth}{
l c
>{\hsize=1.25\hsize\centering\arraybackslash}X
>{\hsize=1.25\hsize\centering\arraybackslash}X
>{\hsize=1.25\hsize\centering\arraybackslash}X
!{\color{black}\vrule width 0.45pt}
>{\hsize=0.85\hsize\centering\arraybackslash}X
>{\hsize=0.85\hsize\centering\arraybackslash}X
>{\hsize=0.85\hsize\centering\arraybackslash}X
!{\color{black}\vrule width 0.45pt}
>{\hsize=0.70\hsize\centering\arraybackslash}X
}
\toprule
\textbf{Model} & \textbf{Size} 
& \textbf{MMSI$^\ast$} 
& \textbf{ViewSpatial$^\ast$} 
& \textbf{MindCube$^{\ast,\ddagger}$} 
& \textbf{3DSR$^\diamond$} 
& \textbf{BLINK$^\diamond$} 
& \textbf{CV$^\diamond$} 
& \textbf{Avg.} \\
\midrule

RandomChoice & -- & 25.0 & 26.3 & 33.0 & 45.8 & 38.1 & 42.5 & 35.1 \\
\specialrule{\lightrulewidth}{\aboverulesep}{0pt}

\rowcolor{black!5}\multicolumn{9}{l}{\rule{0pt}{2.3ex}\textit{Close-source models}}\\
Gemini-2.5-Pro & -- & 38.0 & 46.0 & 57.6 & 59.3 & 73.5 & 85.9 & 60.1 \\
GPT-5 & -- & 41.8 & 45.5 & 56.3 & 60.3 & 68.0 & 84.6 & 59.4 \\
Gemini-3-Pro-Preview & -- & 45.2 & 50.3 & 70.8 & 68.9 & 76.0 & 92.0 & 67.2 \\
\specialrule{\lightrulewidth}{\aboverulesep}{0pt}

\rowcolor{black!5}\multicolumn{9}{l}{\rule{0pt}{2.3ex}\textit{Open-source General Models}}\\
InternVL3-2B~\citep{zhu2025internvl3} & 2B & 26.5 & 32.5 & 37.5 & 47.7 & 50.8 & 76.5 & 45.3 \\
InternVL3-8B~\citep{zhu2025internvl3} & 8B & 28.0 & 38.6 & 41.5 & 44.3 & 53.5 & 81.0 & 47.8 \\
Qwen2.5-VL-3B-Instruct~\citep{bai2025qwen2} & 3B & 28.6 & 31.9 & 37.6 & 43.5 & 48.7 & 71.8 & 43.7 \\
Qwen2.5-VL-7B-Instruct~\citep{bai2025qwen2} & 7B & 26.8 & 36.8 & 36.0 & 47.5 & 55.9 & 75.4 & 46.4 \\
Qwen3-VL-2B-Instruct~\citep{bai2025qwen3vltechnicalreport} & 2B & 28.9 & 36.9 & 34.5 & 47.5 & 53.2 & 78.4 & 46.6 \\
Qwen3-VL-4B-Instruct~\citep{bai2025qwen3vltechnicalreport} & 4B & 30.1 & 42.5 & 37.0 & 52.5 & 62.6 & 84.7 & 51.6 \\
Qwen3-VL-8B-Instruct~\citep{bai2025qwen3vltechnicalreport} & 8B & 31.0 & 42.2 & 29.4 & 53.9 & 66.7 & 85.1 & 51.4 \\

\specialrule{\lightrulewidth}{\aboverulesep}{0pt}

\rowcolor{black!5}\multicolumn{9}{l}{\rule{0pt}{2.3ex}\textit{Spatial Intelligence Models}}\\
SpatialLadder-3B~\citep{li2025spatialladder} & 3B & 27.4 & 39.8 & 43.4 & 42.8 & 43.0 & 73.7 & 45.0 \\
% Spatial-MLLM-4B~\citep{wu2025spatial} & 4B & 26.1 & 34.6 & 33.4 & 36.2 & 40.5 & -- & 34.2 \\
SpaceR-7B~\citep{ouyang2025spacer} & 7B & 27.4 & 35.8 & 37.9 & 40.5 & 49.6 & 74.8 & 44.3 \\
ViLaSR-7B~\citep{wu2025reinforcing} & 7B & 30.2 & 35.7 & 35.1 & 46.6 & 51.4 & 76.7 & 46.0 \\
Cambrian-S-3B~\citep{yang2025cambrian} & 3B & 25.2 & 39.0 & 32.5 & 41.4 & 37.7 & 75.2 & 41.8 \\
Cambrian-S-7B~\citep{yang2025cambrian} & 7B & 25.8 & 40.9 & 39.6 & 45.0 & 37.9 & 76.9 & 44.4 \\
VST-3B-RL~\citep{yang2025visual} & 3B & 32.0 & 45.0 & 36.4 & 56.5 & 57.2 & 84.2 & 51.9 \\
VST-7B-RL~\citep{yang2025visual} & 7B & 34.8 & 42.4 & 39.1 & 60.1 & 62.6 & 86.5 & 54.3 \\
Vanilla GRPO % \rule{0pt}{2.3ex} 
& 8B & 32.2 & 44.4 & 44.7 & 59.4 & 65.7 & 85.2 & 55.3 \\

\specialrule{\lightrulewidth}{\aboverulesep}{0pt}

\rowcolor{blue!2}\multicolumn{9}{l}{\rule{0pt}{2.3ex}\textit{Ours}}\\

\rowcolor{blue!10} \method \rule{0pt}{2.3ex} 
& 8B & \textbf{36.1} & \textbf{46.9} & \textbf{48.4} 
& 60.1 & 66.8 & 85.5 & \textbf{57.3} \\

\specialrule{\heavyrulewidth}{0pt}{\belowrulesep}
\end{tabularx}
\caption{Overall accuracy (\%) on six evaluation benchmarks. Benchmarks marked with $\ast$ follow the same multi-view spatial reasoning domain as our training data, while benchmarks marked with $\diamond$ are used as out-of-domain evaluations. MindCube is additionally marked with $\ddagger$ because we only use it for evaluation and do not use its provided training set.}
\label{tab:main_results}
\end{table*}

%% file: tabs/result_qwen3.tex
\begin{table}[t]
\centering
\scriptsize
\setlength{\tabcolsep}{4pt}
\renewcommand{\arraystretch}{1.15}
\resizebox{\linewidth}{!}{%
\begin{tabular}{l c c c c c c c c c c c c}
\toprule
\multirow{2}{*}{\textbf{Models}} &
\multicolumn{6}{c}{\textbf{Positional Relationship}} &
\multicolumn{2}{c}{\textbf{Attribute}} &
\multicolumn{2}{c}{\textbf{Motion}} &
\textbf{MSR} &
\textbf{Avg.} \\
\cmidrule(lr){2-7}\cmidrule(lr){8-9}\cmidrule(lr){10-11}
& Cam.--Cam. & Obj.--Obj. & Reg.--Reg. & Cam.--Obj. & Obj.--Reg. & Cam.--Reg.
& Meas. & Appr.
& Cam. & Obj.
& -- & \\
\midrule
RandomChoice & 25.0 & 25.0 & 25.0 & 25.0 & 25.0 & 25.0 & 25.0 & 25.0 & 25.0 & 25.0 & 25.0 & 25.0 \\
\specialrule{\lightrulewidth}{\aboverulesep}{0pt}

\rowcolor{black!5}\multicolumn{13}{l}{\rule{0pt}{2.3ex}\textit{Close-source models}}\\
Gemini-2.5-Pro & 38.7 & 34.0 & 40.7 & 44.2 & 38.8 & 41.0 & 62.5 & 30.3 & 39.2 & 25.0 & 33.3 & 38.0 \\
GPT-5 & 41.9 & 33.0 & 35.8 & 49.8 & 42.4 & 68.7 & 54.7 & 37.4 & 28.3 & 40.8 & 36.4 & 41.8 \\
\specialrule{\lightrulewidth}{\aboverulesep}{0pt}

\rowcolor{black!5}\multicolumn{13}{l}{\rule{0pt}{2.3ex}\textit{Open-source General Models}}\\
InternVL3-2B~\citep{zhu2025internvl3} & 31.2 & 22.3 & 28.4 & 30.2 & 28.2 & 28.9 & 25.0 & 22.7 & 16.2 & 28.9 & 26.8 & 26.5 \\
InternVL3-8B~\citep{zhu2025internvl3} & 22.6 & 22.3 & 34.6 & 31.4 & 42.4 & 33.7 & 25.0 & 19.7 & 20.3 & 34.2 & 24.8 & 28.0 \\
Qwen2.5-VL-3B-Instruct~\citep{bai2025qwen2} & 36.6 & 30.9 & 28.4 & 26.7 & 28.2 & 31.3 & 31.2 & 16.7 & 16.2 & 35.5 & 28.8 & 28.6 \\
Qwen2.5-VL-7B-Instruct~\citep{bai2025qwen2} & 28.0 & 26.6 & 19.8 & 32.6 & 38.8 & 28.9 & 23.4 & 21.2 & 20.3 & 30.3 & 24.8 & 26.8 \\
Qwen3-VL-2B-Instruct~\citep{bai2025qwen3vltechnicalreport} & 26.9 & 29.8 & 30.9 & 38.4 & 35.3 & 33.7 & 23.4 & 28.8 & 29.7 & 28.9 & 21.2 & 28.9 \\
Qwen3-VL-8B-Instruct~\citep{bai2025qwen3vltechnicalreport} & 28.0 & 37.2 & 32.1 & 31.4 & 35.3 & 38.5 & 37.5 & 15.2 & 27.0 & 28.9 & 29.8 & 31.1 \\

\specialrule{\lightrulewidth}{\aboverulesep}{0pt}

\rowcolor{black!5}\multicolumn{13}{l}{\rule{0pt}{2.3ex}\textit{Spatial Intelligence Models}}\\
SpatialLadder-3B~\citep{li2025spatialladder} & 36.6 & 29.8 & 29.6 & 32.6 & 30.6 & 24.1 & 18.8 & 31.8 & 23.0 & 23.7 & 23.2 & 27.4 \\
Spatial-MLLM-4B~\citep{wu2025spatial} & 24.7 & 21.3 & 28.4 & 30.2 & 29.4 & 28.9 & 18.8 & 34.9 & 10.8 & 23.7 & 29.8 & 26.1 \\
SpaceR-7B~\citep{ouyang2025spacer} & 25.8 & 31.9 & 29.6 & 25.6 & 31.8 & 22.9 & 26.6 & 28.8 & 16.2 & 34.2 & 27.3 & 27.4 \\
ViLaSR-7B~\citep{wu2025reinforcing} & 29.0 & 35.1 & 28.4 & 39.5 & 40.0 & 44.6 & 31.2 & 16.7 & 17.6 & 31.6 & 23.2 & 30.2 \\
Cambrian-S-3B~\citep{yang2025cambrian} & 25.8 & 28.7 & 24.7 & 48.8 & 24.7 & 33.7 & 29.7 & 22.7 & 20.3 & 28.9 & 18.7 & 27.0 \\
Cambrian-S-7B~\citep{yang2025cambrian} & 24.7 & 26.6 & 24.7 & 47.7 & 22.4 & 31.3 & 32.8 & 24.2 & 12.2 & 30.3 & 24.2 & 27.1 \\
Vanilla GRPO \rule{0pt}{2.3ex} & 47.3 & 29.8 & 28.4 & 34.9 & 32.9 & 43.4 & 43.8 & 27.3 & 32.4 & 27.6 & 21.2 & 32.2 \\

\specialrule{\lightrulewidth}{\aboverulesep}{0pt}

\rowcolor{blue!2}\multicolumn{13}{l}{\rule{0pt}{2.3ex}\textit{Ours}}\\

\rowcolor{blue!10} \method \rule{0pt}{2.3ex} & \textbf{48.4} & \textbf{35.1} & \textbf{29.6} & \textbf{48.8} & \textbf{40.0} & \textbf{55.4} & \textbf{50.0} & 19.7 & \textbf{33.8} & 23.7 & \textbf{24.8} & \textbf{36.1} \\[-0.2ex]

\specialrule{\heavyrulewidth}{0pt}{0pt}
\end{tabular}
}
\caption{Fine-grained accuracy (\%) breakdown on MMSI-Bench. We compare \method with baselines that are included in the main results table and also have fine-grained MMSI results available.}
\label{tab:finegrained_mmsi}
\end{table}

%% file: tabs/ablation.tex
\begin{table}[t]
\centering
\scriptsize
\setlength{\tabcolsep}{4pt}
\renewcommand{\arraystretch}{1.15}
\resizebox{\linewidth}{!}{%
\begin{tabular}{l c c c c c c c c c c c c}
\toprule
\multirow{2}{*}{\textbf{Models}} &
\multicolumn{6}{c}{\textbf{Positional Relationship}} &
\multicolumn{2}{c}{\textbf{Attribute}} &
\multicolumn{2}{c}{\textbf{Motion}} &
\textbf{MSR} & 
\textbf{Avg.} \\
\cmidrule(lr){2-7}\cmidrule(lr){8-9}\cmidrule(lr){10-11}
& Cam.--Cam. & Obj.--Obj. & Reg.--Reg. & Cam.--Obj. & Obj.--Reg. & Cam.--Reg.
& Meas. & Appr.
& Cam. & Obj.
& -- & \\
\midrule
Qwen3-VL-8B-Instruct & 28.0 & 37.2 & 32.1 & 31.4 & 35.3 & 38.5 & 37.5 & 15.2 & 27.0 & 28.9 & 29.8 & 31.1 \\
Vanilla GRPO & 47.3 & 29.8 & 28.4 & 34.9 & 32.9 & 43.4 & 43.8 & 27.3 & 32.4 & 27.6 & 21.2 & 32.2 \\
w/o Staged Length Reward & 45.2 & 34.0 & 23.5 & 46.5 & 32.9 & 47.0 & 45.3 & 25.8 & 27.0 & 26.3 & 24.8 & 33.5 \\
Full Method & 48.4 & 35.1 & 29.6 & 48.8 & 40.0 & 55.4 & 50.0 & 19.7 & 33.8 & 23.7 & 24.8 & 36.1 \\
\bottomrule
\end{tabular}
}
\caption{Ablation study on MMSI-Bench (accuracy, \%), comparing vanilla GRPO, the variant without the staged length reward, and the full method on fine-grained subcategories and overall performance.}
\label{tab:ablation}
\end{table}

%% file: 5conclusion.tex
\section{Conclusion}

In this paper, we propose \method, a GRPO-based training framework for improving multi-view spatial reasoning through guided-prompt consistency and staged length regulation. By comparing model behavior under standard and guided prompts, our method provides dense trajectory-level feedback without introducing any inference-time overhead, while the staged length guard encourages reasoning that is sufficient but not unnecessarily verbose. Experiments on six benchmarks show that \method consistently improves over the base Qwen3-VL-8B-Instruct model and vanilla GRPO, with especially clear gains on in-domain multi-view spatial reasoning tasks. Fine-grained results further indicate that the improvements mainly come from stronger cross-view positional reasoning. Overall, \method demonstrates that explicitly regulating reasoning behavior is an effective strategy for building more accurate, compact, and robust multimodal spatial reasoning models.